# Evaluation of Regularization-based Continual Learning Approaches: Application to HAR

Bonpagna Kann, Sandra Castellanos-Paez, Philippe Lalanda
*Université Grenoble Alpes*
*Laboratoire d'Informatique de Grenoble*
Grenoble, France
firstname.lastname@univ-grenoble-alpes.fr

*Abstract*— Pervasive computing allows the provision of services in many important areas, including the relevant and dynamic field of health and well-being. In this domain, Human Activity Recognition (HAR) has gained a lot of attention in recent years. Current solutions rely on Machine Learning (ML) models and achieve impressive results. However, the evolution of these models remains difficult, as long as a complete re-training is not performed. To overcome this problem, the concept of Continual Learning is very promising today and, more particularly, the techniques based on regularization. These techniques are particularly interesting for their simplicity and their low cost. Initial studies have been conducted and have shown promising outcomes. However, they remain very specific and difficult to compare. In this paper, we provide a comprehensive comparison of three regularization-based methods that we adapted to the HAR domain, highlighting their strengths and limitations. Our experiments were conducted on the UCI HAR dataset and the results showed that no single technique outperformed all others in all scenarios considered.

*Keywords*— Continual learning, regularization methods, HAR

## I. INTRODUCTION

Pervasive computing has imposed itself in our lives for several years now and allows the provision of services in many important areas such as building management, industrial production, or driving and transport assistance [1]. A particularly relevant and dynamic field is that of health and well-being. There, the use of intelligent sensors worn or inserted in the living environments allows to collect relevant information, to conduct analysis, to give advice, and even to act on some physiological parameters if needed. Human Activity Recognition (HAR) belongs to this latter domain. Its purpose is to automatically determine basic activities of a person (like standing, walking, running, etc.) which, if contextualized, allow to infer more complex activities (like dining, cooking, playing, etc.). The most popular approach today to implement an HAR application, because it is not very intrusive as opposed to cameras, is the use of wearable devices such as smartphones or smart watches. Data from accelerometers and gyroscopes is usually used to feed a (deep) machine learning model (ML). Results obtained today, mainly in controlled conditions, are of high quality and the state of the art is regularly improved.

An important challenge, however, is that models must be adapted to the characteristics of the clients (how they walk, run, stand, etc.). This need for adaptation is not limited to the initial deployment phase of the models, but continues throughout their life cycle. This is due to the physical evolution of the subjects, who for example can be affected by a disease or age [2]. Updating a model with new data is however a difficult exercise, especially when the additional data is non IID (Independent and Identically Distributed).

If it is fine-tuned sequentially over time, a model tends to give excessive weight to new data, and progressively forget older data. This problem of not being able to recognize initially learned labels when new labels are integrated in the training process is known as "Catastrophic Forgetting" [3]. An existing solution to preserve old knowledge is to regularly perform complete retraining with all the data, old and new. This approach is expensive and risky. It requires important computational and storage resources, and is time consuming. It also requires personal, sensitive information to be sent up to a cloud since the devices generally do not have the necessary capabilities. This clearly raises security and privacy concerns.

An alternative method to deals with catastrophic forgetting is called Continual Learning (CL) [4,5,6]. This approach seeks to retain the accuracy of a ML model in class-incremental scenarios where new classes are sequentially added for training. Here, we focus on a family of CL algorithms known as regularization-based because they are simple and compatible with limited computational resources. These methods use previous version of a model to maintain accuracy on all classes. Specifically, they add terms to the loss function that penalize the model for forgetting earlier knowledge. Initial studies have been conducted and have shown promising outcomes [7,8]. However, they remain specific and difficult to compare.

In this paper, we evaluate three regularization methods to analyze their strengths and weaknesses on the HAR domain. In Section 2 we present the context of HAR and the related work regarding continual learning. In Section 3, we describe the different methods and the methodology used to evaluate them. Section 4 presents our results and, finally, we conclude in Section 5 and sketch our future work

## II. RELATED WORK

Continual Learning (CL), also called lifelong learning or online machine learning, is the ability of a model to learn continually from an infinite stream of data, gradually integrating new acquired knowledge into old knowledge [9]. Going back to our HAR use case, the idea behind CL is to be able to learn new activities or new ways to perform activities (data drift). Previous studies primarily focused on the adaptation of the model to the continuously changing activities of people. In [10], Gjoreski and Roggen used the agglomerative clustering technique to cluster the real-time sensor data. With the use of temporal constraints, the new incoming data will be clustered as an anomaly if it does not belong to any existing clusters. However, this method can lead to many unnecessary clusters resulting from the constant updates of new activities from users. To address this issue, Cheng et al. introduced in [11], the use of the knowledge-driven model, which contains the semantic relationship between user activities and sensor attributes from accelerometer data, to recognize a new activity in the case of small-scale training data. After a new activity is added to the training, the new attributes and their relationship to the existing activities will be recorded by the domain experts manually. However, this method requires a great re-engineering effort as the core process is conducted manually by the domain specialists. On top of that, the ML models also need to be rebuilt or re-trained from scratch when extending the new activities.

More generally, continual learning methods have been categorized into three main families [12]: regularization-based methods, replay methods, and parameter isolation methods. Replay methods store samples from each training task in either raw format or pseudo-samples to be used as the inputs in the next training task, which is essential to avoid forgetting when training on a new task. iCarl [13] stores exemplars, i.e., a subset of samples with the best approximate means of each class, to represent the information from the previous classes. The exemplars are then used as additional inputs to the new training task to retain the good accuracy on the classes from previous tasks. To overcome the problem of overfitting that iCarl is prone to, GEM [14] proposed to constrain the updates of the new task without affecting the previous tasks through the estimated gradient direction projection on the feasible region using first-order Taylor series approximation. However, storing the sample data requires large storage space, and affects user privacy.

Parameter isolation methods create or freeze different parameters for each model training to avoid forgetting. For example, when the new tasks are different from the previous tasks, new neuron branches are created in the network, and the parameters from the previous tasks are frozen, making task-specific components of the network [15]. However, parameter isolation methods have the problem of scalability, leading the complexity of the architecture to grow larger along with the increasing number of learned tasks.

Regularization-based methods use regularization term in the loss function to avoid forgetting. The aim is to retain the knowledge from the previous training when adding new data without having to modify the network architecture. Li and Hoiem proposed Learning Without Forgetting (LwF) [16]. This uses the probabilistic output from the previous model to transfer the knowledge to the new task. This method depends heavily on the relevance of the training tasks as it uses the direct output from the previous task to transfer knowledge. To alleviate this issue, [17] proposed Elastic Weight Consolidation (EWC). EWC estimates the importance of the parameters using the posterior distribution with a Bayesian approach. It penalizes the model when there is a significant change to those important parameters in the new task. Despite this specialization of EWC, this algorithm has shown drawbacks when learning new classes incrementally as shown in [18]. As EWC and LwF work in different aspects of the model training, Kim et al. proposed in [19] the combined model of EWC and LwF.

### III. EVALUATION

We have implemented the three techniques mentioned here before and applied them to our HAR use case. To assess the results, we have followed the evaluation of class-incremental methodology where we begin by performing two-class classification, and add one different class in each round. In our study, an activity is referred to as a class while a collection of one or more activities added each round is described as a task. We implement 6 rounds of local training for the evaluation of LwF, EWC, and EWCLwF.

#### A. Regularization Methods

LwF uses knowledge distillation (KD) loss as a regularization term, transferring the knowledge from teacher to student model. Distillation loss $L_{KD}$ is described by (1), where $l$ is defined as the number of classes, $y_0^{\prime(i)}$ the prediction logits of the teacher model, $\hat{y}_0^{\prime(i)}$ the prediction logits of the student model for an old class label $i$.

$$L_{KD}(y_0, \hat{y}_0) = -\sum_{i=1}^{l} y_o^{\prime(i)} \log \hat{y}_0^{\prime}(i) \quad (1)$$

The temperature-scaled logits for both teacher and student model are computed as shown in (2), where T is the temperature scaling parameter.

$$y_0^{(i)} = \frac{(y_0^{(i)})^{\frac{1}{T}}}{\sum_j (y_o^{(j)})^{\frac{1}{T}}} \text{ and } \hat{y}_0^{(i)} = \frac{(\hat{y}_0^{(i)})^{\frac{1}{T}}}{\sum_j (\hat{y}_0^{(j)})^{\frac{1}{T}}} \quad (2)$$

With the use of the softmax cross-entropy for the classification loss $L_B$ from the student model, the final loss in LwF is the combination of the classification loss and the knowledge distillation loss denoted as (3):

$$L = \alpha L_B + (1 - \alpha) L_{KD} \quad (3)$$

where α is a scalar which regularizes the influence of each term.

EWC estimates the importance of the model parameters $\theta$ with the posterior distribution $\log p(\theta|D)$ of the parameter given the whole dataset illustrated in (4), where $\log p(D_B|\theta)$ refers to the conventional loss for task B, $\log p(\theta|D_A)$ the important parameters of task A, and $log\,p(D_B)$ the distribution of the dataset in task B.

$$log\,p(\theta|D) = log\,p(D_B|\theta) + log\,p(\theta|D_A) - log\,p(D_B) \quad (4)$$

However, since the true probability $log\,p(\theta|D_A)$ is intractable, it can be estimated with Laplace approximation as a Gaussian distribution with a diagonal precision matrix determined by the Fisher Information Matrix (FIM). The loss function of EWC is written in (5), where $L_B$ is the loss of the new task B, λ refers to the importance of the old task A to task B, ($\theta_i - \theta^*_{A,i}$) refers to the changes of the important parameters in task A after training on task B, and $F_i$ is the diagonal FIM for each task $i$.

$$L = L_B(\theta) + \sum_i \frac{\lambda}{2} F_i(\theta_i - \theta^*_{A,i})^2 \quad (5)$$

EWCLwF was proposed to combine EWC and LwF. The loss function is denoted in (6):

$$L = L_B(\theta) + L_{LwF}(\theta) + L_{EWC}(\theta) \quad (6)$$

where $L_{LwF}$ is the total loss of the LwF and $L_{EWC}$ is the total loss from the EWC method.

### B. Metrics

To evaluate the model specialization, we use the overall average accuracy, the average accuracy at a specific task, and forgetting [20].

**Overall average accuracy**: The average accuracy of the model on classes which were already learned by the model.

$$A = \frac{1}{R}\sum_{r=1}^{R} a_r \quad (7)$$

where the accuracy $a_r$ is calculated on classes which have been already learned between round 1 and round $r$.

**Average accuracy at task t:** The average accuracy of the model on task d after training on task t.

$$A_t = \frac{1}{t}\sum_{d=1}^{t} a_{t,d} \quad (8)$$

$$a_{t,d} = \frac{1}{r^t}\sum_{r'=r^{t'}}^{r^{t'}+r^t} a_{t,d}^{r'} \quad (9)$$

where $a_{t,d}$ is calculated by taking all the examples of classes corresponding to an old task d from the test set, then calculating the accuracy of the model on them after learning a new task t.

**Forgetting**: The average forgetting of the model on task d after performing the task $t$. The higher $F_t$ is, the more model forgets.

$$F_t = \frac{1}{t-1}\sum_{d=1}^{t-1} f_{t,d} \quad (10)$$

$$f_{t,d} = max_{i\in\{d,\ldots,t-1\}}\{a_{i,d}\} - a_{t,d} \quad (11)$$

### C. Dataset

The UCI HAR dataset contains the data of six activities from 30 volunteers. Each activity contains 128 recordings from the accelerometer and gyroscope for the x, y, z axes. The six activities and their labels in the data are described as follows: Walking (0), Walking Up (1), Walking Down (2), Sitting (3), Standing (4), and Laying (5). The dataset contains 10299 records with 561 feature vectors with time and frequency domain variables.

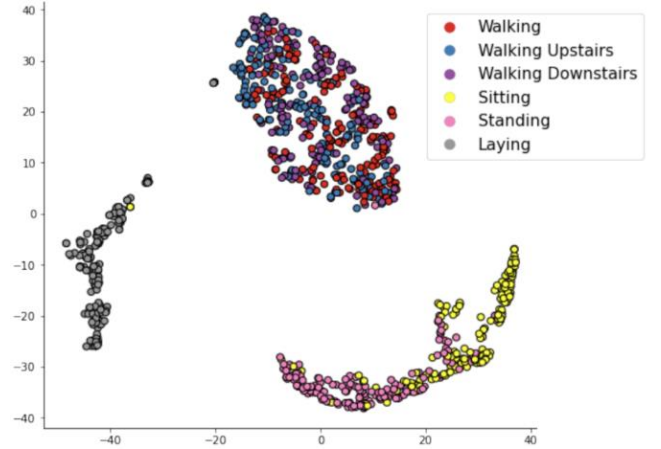

Fig. 1. t-SNE Analysis of UCI HAR Dataset

In order to provide a deeper understanding of the dataset, a t-SNE analysis [21] on the last dense layer before the final activation of a pre-trained CNN model using 960 samples was conducted. In Fig. 1, we present a visualization using 160 random samples from each class. It can be seen that "Laying" is clearly separated from other classes while walking-related activities stay in the same cluster, displaying similar characteristics between activities. Additionally, "Sitting" and "Standing" are also located near each other, forming another cluster in the dataset.

### D. Evaluation Process of Regularization Methods

Python 3 has been used in this experiment along with TensorFlow 2. We used a CNN model architecture from a previous study [7] which includes 196 filters of a 16x1 convolution layer followed by a 1x4 max pooling layer, then 1024 units of a dense layer, and a softmax layer with SGD optimizer. The initial weights for the CNN model were obtained by pretraining the previous study CNN model on a well-balanced dataset of 10 examples per class. We used dropout rate equal of 0.5, and batch size B =32 to both LwF and EWC. The learning rate of [0.001, 0.01], α = 0.1 and T=3 is set for LwF part, while the learning rate of [10$^{-5}$, 5x10$^{-3}$], λ = 5 is defined for EWC part. We trained the model with the class-incremental method for 6 rounds, starting with two classes by learning the same classes with different data for 2-3 rounds before adding another class for the remaining rounds. The size of the training data for each round is 120 for each class.

In order to study how each method performs, the following three scenarios were considered:
- Scenario 0: Learning similar activities before switching to a completely different activity.
- Scenario 1: Learning different types of activities continuously.
- Scenario 2: Learning similar activities continuously.

More information is given in Table I.

TABLE I. TRAINING SCENARIOS OF CL ALGORITHMS

| Methods | Training Classes |
|---|---|
| Scenario 0 (Same cluster, then different clusters) | Case 1: Round 1-3 : 0: Walking, 1: Walking Upstairs; Round 4-6 : 5 : Laying<br>Case 2: Round 1-2 : 1: Walking Upstairs, 2 : Walking Down; Round 3-4 : 4: Standing; Round 5-6 : 5 : Laying |
| Scenario 1 (Different clusters, then different clusters) | Case 1: Round 1-3 : 1: Walking Upstairs, 4: Standing; Round 4-6 : 5 : Laying<br>Case 2: Round 1-3 : 2 : Walking Down, 5 : Laying; Round 4-6 : 3 : Sitting |
| Scenario 2 (Same cluster, then same cluster) | Round 1-3 : 0: Walking, 1: Walking Upstairs, Round 4-6 : 2 : Walking Down |

## IV. RESULTS AND DISCUSSION

### A. Evaluation of Regularization Methods

**Scenario 0:** The results show that EWC has the highest average accuracy and the lowest forgetting in the final round as the class from the new cluster is added to the model after learning the two similar classes as shown in Table II.

TABLE II. AVERAGE ACCURACY (A), ACCURACY BY TASKS ($A_{1-3}$), AND FORGETTING BY TASKS ($F_{2-3}$) OF SCENARIO 0

|   | Case 1 (Class 0, 1, 5) | | | Case 2 (Class 1, 2, 4, 5) | | |
|---|---|---|---|---|---|---|
|   | EWC | LWF | EWCLWF | EWC | LWF | EWCLWF |
| A | 0.863 | 0.807 | 0.836 | 0.625 | 0.607 | 0.618 |
| $A_1$ | 0.831 | 0.724 | 0.861 | 0.586 | 0.582 | 0.580 |
| $A_2$ | 0.895 | 0.800 | 0.811 | 0.538 | 0.500 | 0.528 |
| $F_2$ | 0.102 | 0.215 | 0.187 | 0.110 | 0.202 | 0.184 |
| $A_3$ |   |   |   | 0.752 | 0.738 | 0.746 |
| $F_3$ |   |   |   | 0.278 | 0.313 | 0.354 |

In addition, the accuracy by classes also displays how EWC could retain the previous knowledge better in this situation with the higher accuracy after the class from a different cluster is added, which is shown in Fig. 2.

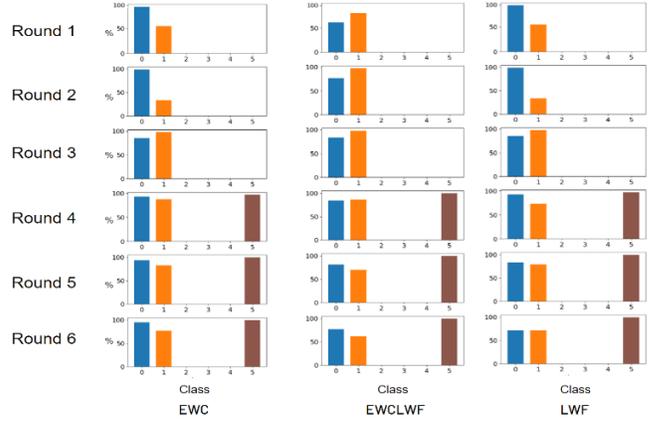

Fig. 2. : The accuracy by class in each round of Case 1 in Scenario 0

**Scenario 1:** According to Table III, after training with two different classes for 3 rounds, another class from another different cluster is added. The results show that EWCLwF has the best performance with the highest average accuracy and the lowest forgetting.

TABLE III. AVERAGE ACCURACY (A), ACCURACY BY TASKS ($A_{1-2}$), AND FORGETTING ($F_2$) OF SCENARIO 1

|   | Case 1 (Class 1, 4, 5) | | | Case 2 (Class 2, 5, 3) | | |
|---|---|---|---|---|---|---|
|   | EWC | LWF | EWCLWF | EWC | LWF | EWCLWF |
| A | 0.884 | 0.864 | 0.895 | 0.813 | 0.801 | 0.828 |
| $A_1$ | 0.888 | 0.861 | 0.875 | 0.885 | 0.878 | 0.881 |
| $A_2$ | 0.855 | 0.866 | 0.875 | 0.881 | 0.881 | 0.776 |
| $F_2$ | 0.281 | 0.305 | 0.228 | 0.265 | 0.285 | 0.237 |

From Fig. 3, with more different classes added to the training, the performance of EWC started to decrease, and so does its forgetting. Consequently, this performance showed a similar outcome as the result in [12], suggesting the weakness of EWC over the long-term as FIM is approximated after optimization of the task, inducing gradients near zero. As a consequence, the regularization strength is initially very high and lowered only to decay stability.

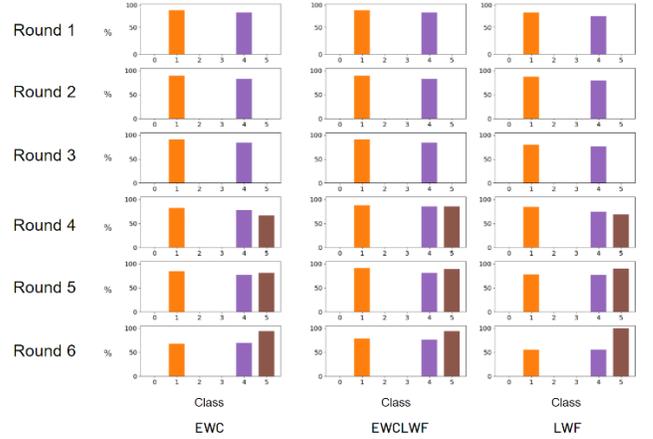

Fig. 3. The accuracy by class in each round of Case 1 in Scenario 1

**Scenario 2:** From Table IV, LwF has outperformed the other two algorithms when learning similar classes. With the support from the teacher model and a direct input-output relationship from the previous training, LwF has shown the highest accuracy while maintaining the lowest forgetting after the training. Additionally, LwF also has better accuracy on each class after a class from the same cluster is added in the new round than the other two regularization-based methods as shown in Fig. 4.

TABLE IV. AVERAGE ACCURACY (A), ACCURACY BY TASKS ($A_{1-2}$), AND FORGETTING ($F_2$) OF SCENARIO 2

|     | Scenario 2 | | |
| --- | --- | --- | --- |
|     | EWC | LwF | EWCLwF |
| A   | 0.527 | 0.558 | 0.546 |
| $A_1$ | 0.651 | 0.706 | 0.663 |
| $A_2$ | 0.431 | 0.410 | 0.428 |
| $F_2$ | 0.379 | 0.215 | 0.357 |

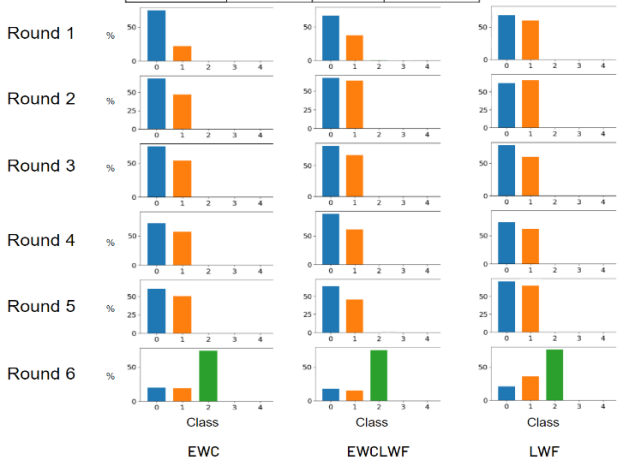

Fig. 4. The accuracy of the model by class in each round of Scenario 2

### B. Discussion

From our experiments, it appears that there is no clear winner for all scenarios. Indeed, when switching activities during learning (Scenario 0), EWC showed the best performances; when learning different types of activities continuously (Scenario 1), EWCLwF performed best; and, when learning similar types of activities continuously (Scenario 2), LwF outperformed the other methods. To explain the difference in performance, we can look at the problem under the plasticity-stability dilemma [1] prism. On the plasticity-stability axis, LwF leans on the stability side since it is a technique that directly penalizes changes in the network; EWC is more on the plasticity side since it tries to find parameters that achieve good compromise between different tasks; and, EWCLwF combining both lies somewhere in the middle. Then, Scenario 0 having a drastic change in activities needs a very plastic method; Scenario 2 however, having only very similar activities is served best by more stable methods; and, Scenario 1 having consistently diverse activities needs a trade-off between plasticity and stability.

---

[1] This dilemma relates to the fact that in order to integrate new knowledge a network needs to be plastic, i.e. be able to change its parameters to learn effectively. However, to retain knowledge, a network should be stable, i.e. be able to keep its parameters compatible with already acquired knowledge.

Our results revealed the problem of no universally good regularization method to effectively perform CL in HAR in various scenarios. Although we experimented in a single dataset, UCI is a balanced dataset known in the HAR community as an entry level challenge. We thus believe that the problem will only be exacerbated on more complex datasets.

One research avenue to this problem could be the combination of regularization and replay methods which can be more all-purpose since the latter have access to some store data. To overcome store space limitations, replay methods could be based on learned data generator as in [22]. Although, learning this generator comes with its own challenges. Another research avenue could be to dynamically switch between several regularization methods. To switch efficiently, one have to automatically identify the scenario in which the learning is performed.

### V. CONCLUSIONS AND FUTURE WORK

In this paper, we evaluated three regularization algorithms on their strengths and weaknesses in different scenarios in the HAR domain. Results have shown that while EWC helped the model coping with classes from different clusters occasionally, EWCLwF improved the weakness of EWC in the long-term single method implementation. On the other hand, LwF was more specialized in training similar classes.

In future work, we would like to investigate a dynamic learning technique that can help the model deal with a variety of classes that the model has not seen before by maintaining accuracy while keeping the forgetting low in class-incremental scenarios. To do so, a research avenue is to study and exploit the strengths and weaknesses of each CL method depending on the scenario. Thus, we could first answer to the problem of coarsely clustering the stream of input classes. Secondly, a meta-learner could adapt and choose which strategy in the portfolio of state-of-the-art CL approaches to use depending on the found clustering.

REFERENCES

[1] C. Becker, C. Julien, P. Lalanda, and F. Zambonelli, "Pervasive computing middleware: Current trends and emerging challenges," *CCF Transactions on Pervasive Computing and Interaction*, vol. 1, no. 1, pp. 10–23, 2019.

[2] Y. J. Xie, E. Y. Liu, E. R. Anson, and Y. Agrawal, "Age-related imbalance is associated with slower walking speed: An analysis from the National Health and Nutrition Examination Survey," *Journal of Geriatric Physical Therapy*, vol. 40, no. 4, pp. 183–189, 2017.

[3] R. M. French, "Catastrophic forgetting in Connectionist Networks," *Encyclopedia of Cognitive Science*, 2006.

[4] F. Wiewel and B. Yang, "Entropy-based sample selection for online continual learning," *2020 28th European Signal Processing Conference (EUSIPCO)*, 2021.

[5] I. Jeon and S. Shin, "Continual representation learning for images with variational continual auto-encoder,"

*Proceedings of the 11th International Conference on Agents and Artificial Intelligence*, 2019.

[6] H. M. Fayek, L. Cavedon, and H. R. Wu, "Progressive learning: A deep learning framework for continual learning," *Neural Networks*, vol. 128, pp. 345–357, 2020.

[7] A. Usmanova, F. Portet, P. Lalanda, and G. Vega, "A distillation-based approach integrating continual learning and federated learning for pervasive services," *arXiv preprint arXiv:2109.04197*, 2021.

[8] S. Jha, M. Schiemer, F. Zambonelli, and J. Ye, "Continual learning in sensor-based human activity recognition: An empirical benchmark analysis," *Information Sciences*, vol. 575, pp. 1–21, 2021.

[9] Z. Chen and B. Liu, *Lifelong machine learning*, 2nd ed. San Rafael, CA: Morgan & Claypool Publishers, 2018.

[10] H. Gjoreski and D. Roggen, "Unsupervised online activity discovery using temporal behaviour assumption," *Proceedings of the 2017 ACM International Symposium on Wearable Computers*, 2017.

[11] H. T. Cheng, F. T. Sun, M. Griss, P. Davis, J. Li, and D. You, "Nuactiv," Proceeding of the 11th annual international conference on Mobile systems, applications, and services, 2013.

[12] M. Delange, R. Aljundi, M. Masana, S. Parisot, X. Jia, A. Leonardis, G. Slabaugh, and T. Tuytelaars, "A continual learning survey: Defying forgetting in classification tasks," *IEEE Transactions on Pattern Analysis and Machine Intelligence*, pp. 1–1, 2021.

[13] S. A. Rebuffi, A. Kolesnikov, G. Sperl, and C. H. Lampert, "ICaRL: Incremental classifier and representation learning," *2017 IEEE Conference on Computer Vision and Pattern Recognition (CVPR)*, 2017.

[14] D. Lopez-Paz and M. A. Ranzato, "Gradient episodic memory for continual learning," in *Proceedings of the 31st International Conference on Neural Information Processing Systems (NIPS'17)*, 2017, pp. 6470–6479.

[15] A. Rakaraddi, L. Siew Kei, M. Pratama, and M. de Carvalho, "Reinforced continual learning for graphs," *Proceedings of the 31st ACM International Conference on Information & Knowledge Management*, 2022.

[16] Z. Li and D. Hoiem, "Learning without forgetting," *IEEE Transactions on Pattern Analysis and Machine Intelligence*, vol. 40, no. 12, pp. 2935–2947, 2018.

[17] J. Kirkpatrick, R. Pascanu, N. Rabinowitz, J. Veness, G. Desjardins, A. A. Rusu, K. Milan, J. Quan, T. Ramalho, A. Grabska-Barwinska, D. Hassabis, C. Clopath, D. Kumaran, and R. Hadsell, "Overcoming catastrophic forgetting in Neural Networks," *Proceedings of the National Academy of Sciences*, vol. 114, no. 13, pp. 3521–3526, 2017.

[18] R. Kemker, M. McClure, A. Abitino, T. Hayes, and C. Kanan, "Measuring catastrophic forgetting in neural networks," *Proceedings of the AAAI Conference on Artificial Intelligence*, vol. 32, no. 1, 2018.

[19] H. E. Kim, S. Kim, and J. Lee, "Keep and learn: Continual learning by constraining the latent space for knowledge preservation in Neural Networks," *Medical Image Computing and Computer Assisted Intervention – MICCAI 2018*, pp. 520–528, 2018. \

[20] M. Masana, X. Liu, B. Twardowski, M. Menta, A. D. Bagdanov, and J. van de Weijer, "Class-incremental learning: Survey and performance evaluation on Image Classification," *IEEE Transactions on Pattern Analysis and Machine Intelligence*, pp. 1–20, 2022.

[21] L. van der Maaten and G. Hinton, "Visualizing data using t-SNE," *Journal of Machine Learning Research*, vol. 9, pp. 2579–2605, 2008.

[22] H. Shin, J. K. Lee, J. Kim, and J. Kim, "Continual learning with deep generative replay," in *31st International Conference on Neural Information Processing Systems (NIPS'17)*, 2017, pp. 2994–3003.